\pdfoutput=1
\documentclass[letterpaper]{article} % DO NOT CHANGE THIS
\usepackage{aaai24}  % DO NOT CHANGE THIS
\usepackage{times}  % DO NOT CHANGE THIS
\usepackage{helvet}  % DO NOT CHANGE THIS
\usepackage{courier}  % DO NOT CHANGE THIS
\usepackage[hyphens]{url}  % DO NOT CHANGE THIS
\usepackage{graphicx} % DO NOT CHANGE THIS
\usepackage{natbib}  % DO NOT CHANGE THIS AND DO NOT ADD ANY OPTIONS TO IT
\usepackage{caption} % DO NOT CHANGE THIS AND DO NOT ADD ANY OPTIONS TO IT
\usepackage{algorithm}

\usepackage{newfloat}
\usepackage{listings}
\DeclareCaptionStyle{ruled}{labelfont=normalfont,labelsep=colon,strut=off} % DO NOT CHANGE THIS
\floatstyle{ruled}
\newfloat{listing}{tb}{lst}{}
\floatname{listing}{Listing}
\usepackage{graphicx}%
\usepackage{subcaption}%
\usepackage{multirow}%
\usepackage{amsmath,amssymb,amsfonts}%
\usepackage{amsthm}%
\usepackage{mathrsfs}%
\usepackage[title]{appendix}%
\usepackage{xcolor}%
\usepackage{textcomp}%
\usepackage{manyfoot}%
\usepackage{booktabs}%
\usepackage{algorithm}%
\usepackage{algorithmicx}%
\usepackage{algpseudocode}%
\usepackage{listings}%
\usepackage{bm}

\title{Semi-parametric Expert Bayesian Network Learning \\ with Gaussian Processes and Horseshoe Priors}
\author{
      Yidou Weng\textsuperscript{\rm 1},
    Finale Doshi-Velez\textsuperscript{\rm 2}
}
\affiliations{
    \textsuperscript{\rm 1}National University of Singapore\\
    \textsuperscript{\rm 2}Harvard University\\
    yidouw@u.nus.edu,
    finale@seas.harvard.edu

}

\begin{document}

\maketitle
%%=============================================================%%
% Prefix	-> \pfx{Dr}
% GivenName	-> \fnm{Joergen W.}
% Particle	-> \spfx{van der} -> surname prefix
% FamilyName	-> \sur{Ploeg}
% Suffix	-> \sfx{IV}
% NatureName	-> \tanm{Poet Laureate} -> Title after name
% Degrees	-> \dgr{MSc, PhD}
% \author*[1,2]{\pfx{Dr} \fnm{Joergen W.} \spfx{van der} \sur{Ploeg} \sfx{IV} \tanm{Poet Laureate} 
%                 \dgr{MSc, PhD}}\email{iauthor@gmail.com}
% \author{
%     AuthorOne,\equalcontrib\textsuperscript{\rm 1,\rm 2}
%     AuthorTwo,\equalcontrib\textsuperscript{\rm 2}
%     AuthorThree,\textsuperscript{\rm 3}\\
%     AuthorFour,\textsuperscript{\rm 4}
%     AuthorFive \textsuperscript{\rm 5}}
% }
% \affiliations {
%     \textsuperscript{\rm 1}AffiliationOne,\\
%     \textsuperscript{\rm 2}AffiliationTwo,\\
%     \textsuperscript{\rm 3}AffiliationThree,\\
%     \textsuperscript{\rm 4}AffiliationFour,\\
%     \textsuperscript{\rm 5}AffiliationFive\\
%     \{email, email\}@affiliation.com,
%     email@affiliation.com,
%     email@affiliation.com,
%     email@affiliation.com
% }
%%=============================================================%%

\begin{abstract}
This paper proposes a model learning Semi-parametric relationships in an Expert Bayesian Network (SEBN) with linear parameter and structure constraints. We use Gaussian Processes and a Horseshoe prior to introduce minimal nonlinear components. To prioritize modifying the expert graph over adding new edges, we optimize differential Horseshoe scales. In real-world datasets with unknown truth, we generate diverse graphs to accommodate user input, addressing identifiability issues and enhancing interpretability. Evaluation on synthetic and UCI Liver Disorders datasets, using metrics like structural Hamming Distance and test likelihood, demonstrates our models outperform state-of-the-art semi-parametric Bayesian Network model.
\end{abstract}

\section{Introduction}\label{sec:introduction}

% Motivation 
In various disciplines, models have been developed to describe phenomena of interest. Mechanistic models are prevalent in medicine, epidemiology, physics, and other fields. Expert Bayesian Networks (EBNs) serve as powerful tools for representing relationships between multiple variables and supporting decision-making.

% Challenge/Issue/Why we can't do it yet
However, the most common approach—utilizing standard structure learning algorithms for data-driven relationship discovery—has drawbacks. Bayesian networks (BN) are often non-identifiable, and expert knowledge encapsulates the scientific understanding of an entire field. Especially in domains where expert knowledge and model interpretability are crucial, and the goal is to advance scientific understanding, it is vital that a learned network extends and refines current relationships between variables rather than completely rewriting them.

However, EBNs do have limitations: experts may face challenges in specifying edges between variables and defining the parametric form of the relationship along an edge, especially for nonlinear relationships. This becomes a significant restriction when dealing with domains featuring non-linear interactions or where the nature of interactions is unknown.

An approach under exploration is learning BN conditional probability distributions (CPDs) with non-parametric densities
\cite{hofmann1995discovering, ickstadt2011nonparametric, friedman2013gaussian}. Non-parametric models do not assume a specific data distribution, providing more flexibility. However, a limitation is that their representation complexity often grows with the amount of available data \cite{stone1980optimal}. Semi-parametric BN combines the simplicity of parametric assumptions when appropriate with the flexibility of nonparametric models when necessary.

% Contributions/Our work
The objective of our work is to learn semi-parametric additions to an EBN to better explain the data, while minimally modifying the expert graph.
Doing so keeps our model aligned with expert understanding and allows an expert to easily review proposed changes. Specifically, SEBN uses Gaussian processes (GPs) as additional components in EBNs to capture nonlinear components of the model. Additionally, it uses a Horseshoe prior to penalize excessive GP edges, especially those not in the linear EBN.

On a large number of randomly-generated synthetic examples, we demonstrate that SEBN can learn minimal changes and additions to an expert network to maximize predictive performance. We optimized the Horseshoe prior over a spectrum of scales and weights based on validation Structural Hamming Distance and test likelihood. We found that differential Horseshoe priors work the best to penalize especially non-expert edges and maintain the expert structure. SEBN outperformed state-of-the-art Semi-Parametric Bayesian Network (SPBN) \cite{atienza2022semiparametric} in learning the correct edges and predicting unseen data.

Lastly we evaluted SEBN on a real-world dataset, where the ground truth structures are not known. When the single right change is not obvious, we provided diverse reasonable options for experts. All these options outperformed SPBN in held-out likelihood.

\section{Related Work}

\paragraph{Expert Graph As Hard and Soft Constraints}
Expert beliefs can be incorporated into Bayesian Network Structure learning as hard or soft constraints. 

\citet{de2009structure} and \citet{tsamardinos2006max} used expert graphs as hard constraints -- no deletions and/or additions. These works are limited to discrete or linear settings. While we preserve linear expert edges as hard constraints, we allow the addition of nonlinear components. This approach enables capturing nuanced relationship in the data that may be challenging for experts to specify.

Expert beliefs have also been incorporated a priori as soft constraints. \citet{heckerman1995learning} penalized differences between expert graph and posterior structure using a single confidence value for all specified edges. In our model, when penalizing the addition of nonlinear components, we allow for a finer-grained specification of confidence to penalize various edge types differently.

\paragraph{Parametric, Nonparametric and Semiparametric Bayesian Network}
Parametric continuous Bayesian Networks, such as Gaussian Bayesian Networks (GBN), are limited to learning linear dependencies among variables. This restriction becomes significant in domains with non-linear interactions or unknown interaction nature.

When parametric assumptions are not met, the performance of parametric models may suffer. Nonparametric methods, such as splines \cite{imoto2003bayesian}, and Kernel estimators \cite{gunn2002structural}, have been explored. 

A limitation of non-parametric models is that their representation complexity usually growing with the data available \cite{atienza2022hybrid}. Semi-parametric methods, considering some parts as parametric and others as nonparametric, offer greater precision for multidimensional cases \cite{gentle2012computational}.

\citet{boukabour2021semiparametric} and \citet{atienza2022hybrid} proposed Semi-parametric Bayesian Network as SBN and SPBN, respectively, outperforming parametric BN in predicting complicated relationship. However, they restrict nodes to be exclusively either linear or non-parametric, posing a challenge for incorporating expert graphs. Graphs with mostly nonparametric nodes tend to be sparse and significantly differ from linear EBN, making the learned nonparametric relationships hard to interpret. In contrast, our approach introduces flexibility by allowing both linear and non-linear components for each edge. We align the linear component with expert belief, while learning non-linear components to capture nuances potentially missed by the expert. With the learned structure largely consistent with the expert graph, our model facilitates easy interpretation of added non-linear edges.

Another limitation of SBN is that it optimizes linear and non-linear components separately, which not be result in the global optimum. We cotrain both components, outperforming the original two-step learning process.

\paragraph{Nonparametric Method: Gaussian Process}

Nonparametric BN estimate all CPDs with nonparametric methods. Various nonparametric methods have been explored for nonparametric BN, such as Kernel Density Estimation (KDE) models \cite{hofmann1995discovering}, an infinite mixture model \cite{ickstadt2011nonparametric}, the Nadaraya–Watson estimator \cite{nadaraya1964estimating, watson1964smooth, boukabour2021semiparametric}, and Gaussian Processes (GPs) \cite{friedman2013gaussian}. In our work, we use GPs to model nonlinear components.

GPs provide flexible learning of continuous dependencies with advantages over other non-linear models. Firstly, GPs have an amplitude hyperparameter that can be used for threshold testing and regularization. Secondly, as a Bayesian method, GPs incorporate domain knowledge through choice of covariance functions or hyperparameter priors. This facilitates expert-guided learning and posterior probability interpretation for network structures.

Thirdly, GPs learn hyperparameters both locally and globally \cite{piironen2017hyperprior}. Locally, for each parent, the GP model adopts a different set of hyperparameters and priors. Globally, an overarching regularization level for a specific node can be posed by a prior on all its hyperparameters.

\paragraph{Regularizing Gaussian Process with Horseshoe Prior}

To preserve the expert structure and manage model complexity, regularization methods have been explored. Common approaches involve statistical testing to prune edges \cite{boukabour2021semiparametric} and regularization terms on structure complexity or differences, such as BDe score \cite{heckerman1995learning}.

Gaussian Processes, inherently biased toward identifying functional dependencies, can be effectively and efficiently regularized by applying priors over their amplitudes. We opted for the Horseshoe (HS) prior for its heavy tails and capacity to accommodate both zero and large values \cite{ghosh2019model}. The heavy tails of HS effectively deactivate GP with a small amplitude, while GPs with a large amplitude can be interpreted as active.

\section{Background and Notation}
\label{sec:background}

\paragraph{Bayesian Network} 

A Bayesian network is represented by a directed acyclic graph (DAG) \(G\) with nodes \(V = \{1, \ldots, n\}\) and directed edges. This network provides a structure for factorizing the joint probability distribution \(P(x)\) involving \(n\) random variables \(\bold{x} = (x_1, \ldots, x_n)\). \({\theta}\) is the set of parameters, with \({\theta_i}\) denoting the set of parameters for the Conditional Probability Distribution (CPD) of node \(i\).

\paragraph{Parameter Learning}
Assuming a fixed structure for a BN, where the graph's edges and the type of CPD for each node are known, we estimate the parameters for each node's CPD to complete the model using maximum likelihood criterion.

Given \(N\) independent and identically distributed samples \(D = \{\bold{x}^1, \ldots, \bold{x}^N\}\), the likelihood function is defined as:

\[
P(D | \theta, G) = \prod_{j=1}^{N} \prod_{i=1}^{n} P(x_{i}^j | \theta_i, \bold{x}_{\text{Pa}(i)}^j),
\]

where \(\bold{x}_{\text{Pa}(i)}\) is the set of the parents of node i. The global likelihood decomposition property ensures that to maximize \(P(D | \theta, G)\), we can maximize 
\[
P_i(D | \theta, G) = P(X_{i} | \theta_i, X_{\text{Pa}(i)})
\]
independently for each node \(i\).

To prevent overfitting, we use the early-stopping criterion \cite{prechelt2002early}. We randomly split the data \(D\) into two disjoint datasets called the training and validation sets, \(D = D_{\text{train}} \cup D_{\text{val}}\). The parameter optimization on \( D_{\text{train}} \) continues until the point where \(\theta_i\) that improves the likelihood \(P(D_{\text{val}} | \theta, G)\) cannot be found for \(\lambda\) iterations. Here, \(\lambda\) represents the patience parameter. When \(\lambda > 0\), the search is permitted to explore less favorable parameters for up to \(\lambda\) iterations. This approach helps mitigate the risk of being trapped at local maxima.

% param learning decompose to max every node
% strutcure learning exact scor

\paragraph{Structure Learning}
Structure learning methods can be broadly categorized into constraint-based and score-based approaches. Constraint-based methods use conditional independence tests to eliminate and orient edges. Score-based methods, on the other hand, aim to maximize an objective function over different graphs. Score-based methods can be further classified into approximate and exact search methods. Approximate methods, like greedy hill-climbing, iteratively modify a graph's structure to maximize a score function. In contrast, exact algorithms guarantee the return of the highest-scoring DAG by treating structure learning as a constrained combinatorial optimization problem. \cite{kitson2023survey} We use an exact score algorithm.

% point out we max each node, put x and y in context of par and node
\paragraph{Gaussian Process}
We model the non-linear relationship between \(x_i\) and its candidate parents \(\mathbf{x}_{\text{Pa}(i)}\) as a Gaussian Process (GP) \cite{friedman2013gaussian}.

A stochastic process over \(X_{\text{Pa}(i)}\) is a function that assigns to each \(\mathbf{x}_{\text{Pa}(i)} \in X_{\text{Pa}(i)}\) a random variable \(x_i\). The process is said to be a GP if for each finite set of value \(\mathbf{x}_{\text{Pa}(i)}^{1:N}\), the distribution over the corresponding random variables \(x_{i}^{1:N}\) is a multivariate normal distribution, whose mean and covariance functions we specify as  \(E[x_i] = \mu(\mathbf{x}_{\text{Pa}(i)})\) and \(\text{Cov}[x_i,x_i'] = K(\mathbf{x}_{\text{Pa}(i)},\mathbf{x}_{\text{Pa}(i)}')\). As the mean can be considered separately as a linear component, for simplicity and without the loss of generality, we can assume mean of GP is 0. The joint distribution of \(x_{i}^{1:N}\) is therefore:

\begin{multline*}
    P(x_{i}^{1:N} \mid \mathbf{x}_{\text{Pa}(i)}^{1:N}) \propto
    \exp\left(-\frac{1}{2}(x_{i}^{1:N})^T\mathbf{K}_{1:N}^{-1}x_{i}^{1:N}\right)
\end{multline*}

where \(\mathbf{J}_{1:N}\) is the vector of means \((\mu(\mathbf{x}_{\text{Pa}(i)}^1), ..., \mu(\mathbf{x}_{\text{Pa}(i)}^M))\) and \(\mathbf{K}_{1:N}\) is the covariance matrix with the \((p,q)\)-th entry \(K(\mathbf{x}_{\text{Pa}(i)}^p,\mathbf{x}_{\text{Pa}(i)}^q)\).

The covariance between points determines properties like periodicity, smoothness, and amplitude of the learned functions. These aspects of the covariance functions are controlled by its hyperparameters. For example, in the Squared Exponential (SE) function, one of the mostly commonly used covariance functions \cite{williams2006gaussian}:
\[
k(\mathbf{x}_{\text{Pa}(i)}^p, \mathbf{x}_{\text{Pa}(i)}^q) = \sigma^2 \exp\left(-\frac{1}{2}\left(\frac{(x_{\text{Pa}(i)}^p - \mathbf{x}_{\text{Pa}(i)}^q)^2}{l^2}\right)\right)
\]
The hyperparameter \(\sigma^2\) controls the amplitude of variation of the function, and the length scale \(l\) controls the smoothness/wiggleness of the function. Each parent \(j\) of node \(i\) has its own set of hyperparameters \(\sigma_j^2\) and \(l_j\).

In noisy observations, assuming additive independent identically distributed Gaussian noise \(\epsilon\) with variance \(\sigma_n^2\), the prior on observations is:
\[
\quad \text{cov}(X_i) =\mathbf{K} + \sigma_n^2 I.
\]

Integrating the likelihood times the prior, the log marginal likelihood is given by:
\begin{multline*}
\log p(x_{i}^{1:N} \mid \mathbf{x}_{\text{Pa}(i)}^{1:N}) = -\frac{1}{2} \left( (\mathbf{K} +
\sigma_n^2 I)^{-1} \right. \\
+ \left. \log \det(\mathbf{K} + \sigma_n^2 I) + N \log(2\pi)\right)
\end{multline*}

\paragraph{Horseshoe Prior}

The Horseshoe prior is set for the amplitudes \(\sigma^2\) where \(\tau\) is the local scale.

\begin{equation*}
    \sigma^2 \mid \tau \sim \mathcal{N}(0, \tau^2I), \quad \tau \sim \mathcal{C}^+(0,b)
\end{equation*}

Specifically, the logarithm of the probability density function for the amplitudes for $p$ parents with respect to \(\tau\) is given by:

\begin{multline*}
     \log P(\sigma^2 \mid \tau) = \sum_{i=1}^p \log \left( \frac{\tau}{\sqrt{(\sigma_i^2)^2 + \tau^2}} \right)\\
     - \sum_{i=1}^p \log \left( 1 + \frac{\tau^2}{(\sigma_i^2)^2} \right) - \frac{p}{2} \log(2\pi)
\end{multline*}

The value of \(\tau\) determines the level of regularization: with a large \(\tau\), all the variables have very diffuse priors with very little shrinkage towards zero, but letting \(\tau \to 0\) will shrink the amplitude \(\sigma^2\) to zero \cite{piironen2017hyperprior}. We will explore a range of HS scales to produce different regularization and thus varying learned structures.

\section{Model}

We model observed values $x_i$ of each node $i$ with candidate parents \(\mathbf{x}_{\text{Pa}(i)}\) in a Bayesian network as:
\[
\mathbf{x_i} = w \cdot \mathbf{x}_{\text{Pa}(i)} + b +\sum_{x_j\in \mathbf{x}_{\text{Pa}(i)}}f_i(x_j) + \epsilon_i
\]
where $w \cdot \mathbf{x}_{\text{Pa}(i)} + b$ is the linear term, $f_i(\textbf{x})$ is a zero-mean GP with SE covariance function with hyperparameters amplitude $\sigma^2$ and length scale $l$ and $\epsilon_i$ is the noise with assumed known variance $\sigma_n^2$.

\section{Learning}

\paragraph{Parameter Learning}

Using global likelihood decomposition, to maximize \(P(D_{\text{train}} | \theta, G)\), we independently maximize $P_i(D_{\text{train}}| \theta, G) = P(X_{i} | \theta_i, X_{\text{Pa}(i)})$
for each node \(i\).

The posterior likelihood is the sum of probability of $X_{i}$ given $X_{\text{Pa}(i)}$ and of the learned GP amplitude given its Horseshoe prior, in logarithm:

\[
\log P(X_{i}|X_{\text{Pa}(i)}, \tau) = \log P(X_{i}|X_{\text{Pa}(i)}, \sigma^2) + w_{HS}\log P(\sigma^2 | \tau)
\]

where $w_{HS}$ is the weight for the Horseshoe prior term. The prior term is weighted to have a reasonable effect on the likelihood as sample size increases. The parameters to be optimized include the linear coefficients, as well as GP lengthscales and amplitudes. 

We compare two learning modes, either two-step or one-step. In two-step learning, the linear parameters are determined beforehand, either set to ground truth or fitted assuming LGBN. Then they are fixed throughout optimizing GP parameters solely. In one-step learning, linear and GP parameters are optimized together.

\paragraph{Structure Learning}

Parameter learning removes candidate parents with learned GP amplitudes smaller than a defined threshold. In other word, our model considers all possible parents and retains only those deemed significant. This approach eliminates the need to iterate through edge operations in approximate search algorithm. Exact search is further plausible, given that the acyclic constraint imposed by the partial topological order provided by the expert graph substantially reduces the search space.

We implement an exact search in node-ordering space modifying Dynamic Programming (DP). DP breaks down the problem by solving small sub-problems first. Every DAG must have at least one leaf node. A DAG with nodes $X$ can be constructed from a leaf node $X_{\text{leaf}}$ and a sub-DAG with nodes $X - \{X_{\text{leaf}}\}$ \cite{singh2005finding}. The maximum graph score is expressed as a recurrence relation:

\begin{align*}
    \text{score}_{\text{max}}(X) &= \text{score}_{\text{max}}(X - \{X_{\text{leaf}}\})\\
    &\quad+ \text{score}_{\text{max}}(X_{\text{leaf}} | \text{Pa}(X_{\text{leaf}}))
\end{align*}

DP exploits this recurrence relationship, ensuring a guaranteed search for the highest-scoring DAG. DP is feasible for \(n \leq 26\) and is useful for learning moderately sized networks \cite{singh2005finding}. Additionally, to save computational cost, we prune paths that violate the partial topological order provided by experts.

Only nodes last in the partial topological order in the current subgraph can be leaf nodes, as all other nodes can be their parents without creating cycles. This ensures consideration of all possible edges without violating the hard constraint. Note that only expert linear edges, not learned GP edges, are hard constraints. Hence, pruning away insignificant GP edges based on amplitude does not affect the soundness of our modified DP algorithm.

The pseudocode for DP is as follows \cite{singh2005finding}:

\begin{algorithm}
\caption{OptOrd(S) -- Store the score of the best network on S in Cache(S)}
\label{alg:optord}
\begin{algorithmic}[1]
\State $\text{bestscore} \gets \infty$
\If{$S = \varnothing$}
    \State \textbf{return} $0.0$
\EndIf
\For{$x \in S \land {x \in \text{LeafNode}}$}
    \If{$\text{Cached}(S - x)$}
        \State $s \gets \text{Cache}(S - x)$
    \Else
        \State $s \gets \text{OptOrd}(S - x)$
    \EndIf
    \State $s \gets s + \text{BestScore}(S, x)$
    \If{$\text{bestscore} > s$}
        \State $\text{Leaf}(S) \gets x$
        \State $\text{bestscore} \gets s$
    \EndIf
    \State $\text{Cache}(S) \gets \text{bestscore}$
\EndFor
\State \textbf{return} $\text{bestscore}$
\end{algorithmic}
\end{algorithm}

\section{Experimental Setup}

\paragraph*{Synthetic Datasets}

We created two synthetic datasets. Both allow either modifying expert linear edges to contain GP components or adding new GP edges, with the second dataset favoring modification.

\begin{enumerate}
    \item \textbf{Independent-addition Dataset (ID):}
    A dataset with a uniform probability of 0.5 for modification and addition.
    
    \item \textbf{Expert-guided Dataset (ED):}
    A dataset where there's a higher likelihood of observing nonlinearity in edges specified by experts. Specifically, each expert edge has a probability of 0.5 for modification and 0.01 for addition.
\end{enumerate}

\paragraph*{Generate random structure.}
Following the topological order of nodes, for each current node:

\begin{enumerate}
    \item Randomly sample from its ancestors to be its linear parents with a 0.5 probability.
    \item Randomly sample from its ancestors to be its GP parents with corresponding probability as specified in ID or ED case.
\end{enumerate}

\paragraph*{Set up parameters}
\begin{itemize}
    \item The intercept $\beta_0 = 0$
    \item For each linear component coefficient $\beta_j = 1$
    \item For each non-linear component $\gamma_j = \cos{2 \pi X_j}$
    \item The noise variance $\sigma_n^2 \sim 0.01$
\end{itemize}

\paragraph*{Example of a synthetic dataset}
\begin{equation*}
    \mathcal{L} =  \begin{pmatrix}
    0 & 0 & 0 & 0 & 0 \\
    0 & 0 & 0 & 0 & 0 \\
    1 & 1 & 0 & 0 & 0 \\
    1 & 1 & 1 & 0 & 0 \\
    1 & 0 & 0 & 0 & 0 \\
\end{pmatrix}
\end{equation*}

In matrix $\mathcal{L}$, for each row $i$ and for each column $j$ where $j < i$, the ones represent linear edges from $X_j$ to $X_i$. 

\begin{equation*}
    \mathcal{G} =  \begin{pmatrix}
    0 & 0 & 0 & 0 & 0 \\
    0 & 0 & 0 & 0 & 0 \\
    1 & 1 & 0 & 0 & 0 \\
    1 & 0 & 0 & 0 & 0 \\
    1 & 1 & 0 & 1 & 0 \\
\end{pmatrix}
\end{equation*}

In matrix $\mathcal{G}$, for each row $i$ and for each column $j$ where $j < i$, the ones represent non-linear edges from $X_j$ to $X_i$. 

These matrices of linear and GP components reflect a generated BN structure with the following relationships:

% FDV: Make sure the captions for (a), (b) etc. are more informative.
\begin{figure*}[h]
    \begin{minipage}{0.33\textwidth}
        \centering
        \begin{subfigure}{\linewidth}
            \includegraphics[width=\linewidth]{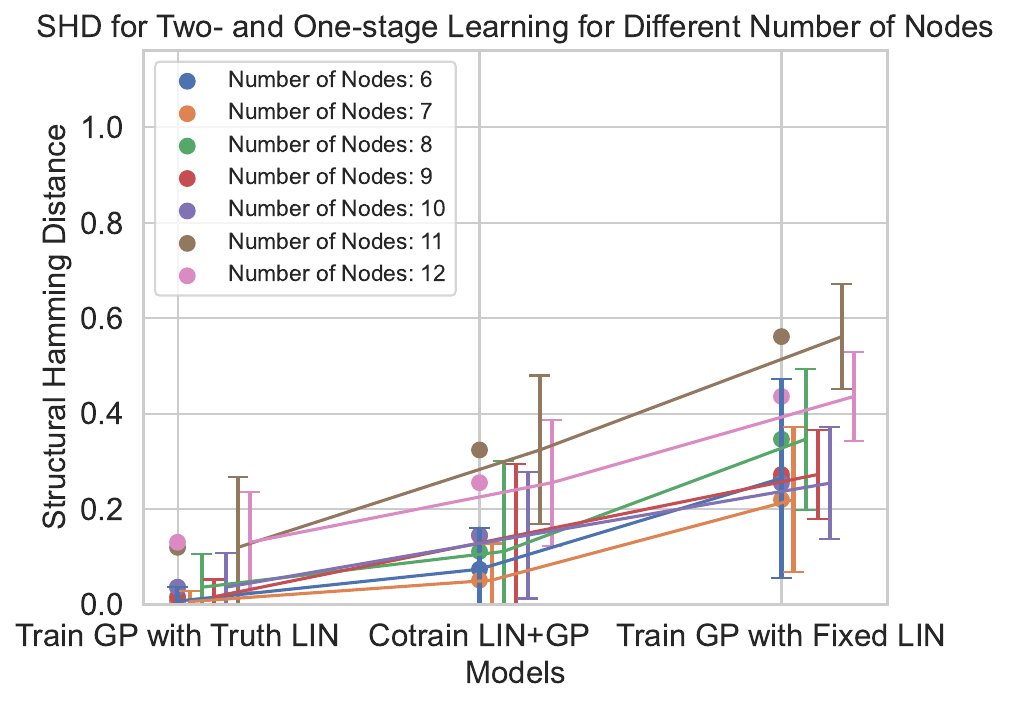}
            \captionsetup{width=0.9\linewidth}
            \caption{ SHD vs. Number Of Nodes for models with ground-truth linear parameters, two-stage trained linear and GP parameters, and co-trained linear and GP parameters. }
            \label{fig:baseline_given_shd}
        \end{subfigure}
    \end{minipage}%
    \begin{minipage}{0.33\textwidth}
        \centering
        \begin{subfigure}{\linewidth}
            \includegraphics[width=\linewidth]{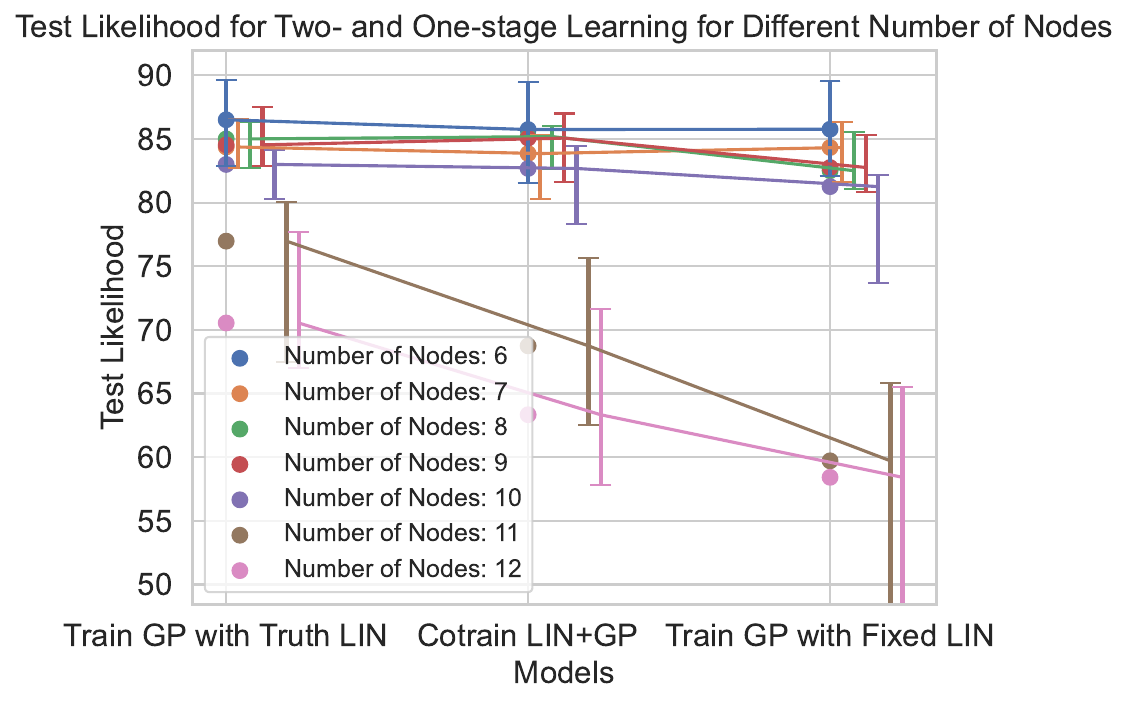}
            \captionsetup{width=0.9\linewidth}
            \caption{Test Likelihood vs. Number Of Nodes for models with ground-truth linear parameters, two-step trained linear and GP parameters, and co-trained linear and GP parameters.}
            \label{fig:baseline:given_lik}
        \end{subfigure}
    \end{minipage}%
    \begin{minipage}{0.33\textwidth}
        \centering
        \begin{subfigure}{\linewidth}
            \includegraphics[width=\linewidth]{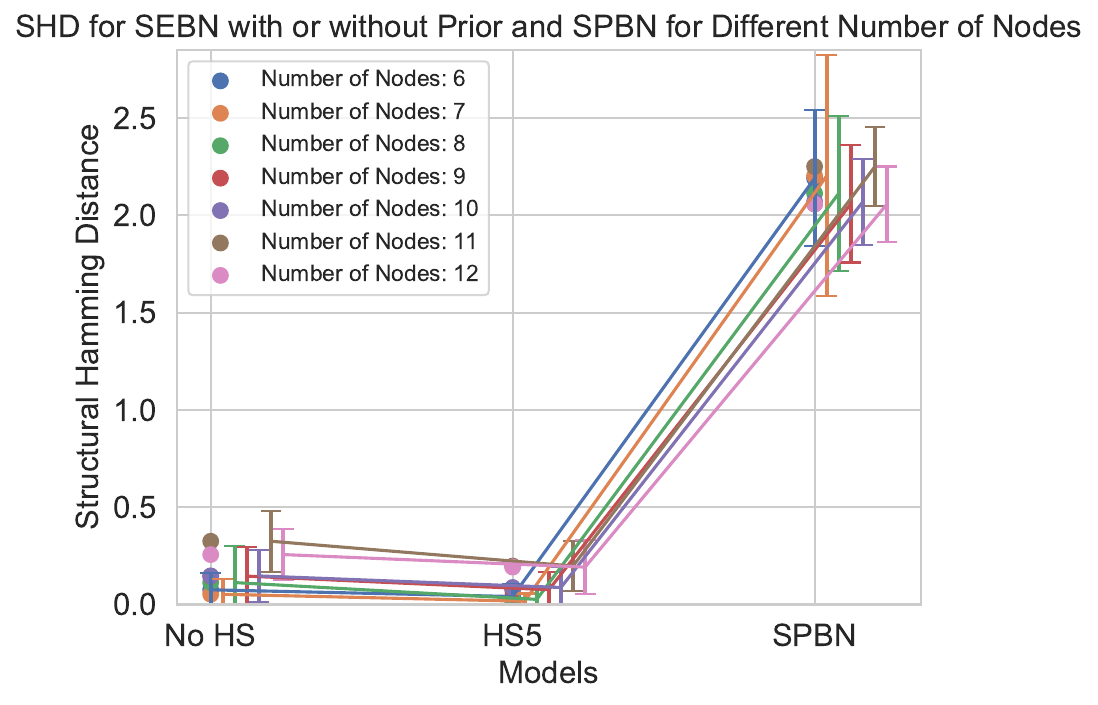}
            \captionsetup{width=0.9\linewidth}
            \caption{SHD vs. Number of Nodes for SPBN baseline and SEBN with no Horseshoe, or with Horseshoe prior scale $\tau = 5$ and weight $w_{HS} = 1$.}
            \label{fig:baseline:spbn}
        \end{subfigure}
    \end{minipage}

    % FDV: What is the take-away?
    % Which data set?
    \caption{In Independent-addition Dataset, (a) and (b) At all network sizes, jointly-training linear and GP parameters got about 50\% closer to the oracle solution than the two-stage approach. (c) SEBN, with or without Horseshoe prior, significantly outperformed the state-of-the-art baseline SPBN}
    \label{fig:baseline}
\end{figure*}
\begin{align*}
    &X_1 \sim N(0,0.1)\\
    &X_2 \sim N(0,0.1)\\
    &X_3 = X_1 + X_2  +\cos(2\pi X_1)  +\cos(2\pi X_2)+ N(0,0.1)\\
    &X_4 = X_1 + X_2 + X_3 + \cos(2\pi X_1) + N(0,0.1)\\
    &X_5 = X_1 +\cos(2\pi X_1) +\cos(2\pi X_2) +\cos(2\pi X_4) + N(0,0.1)
\end{align*}
\begin{figure*}[h]
\centering
    \begin{minipage}{0.4\textwidth}
        \centering
        \begin{subfigure}{\linewidth}
            \includegraphics[width=\linewidth]{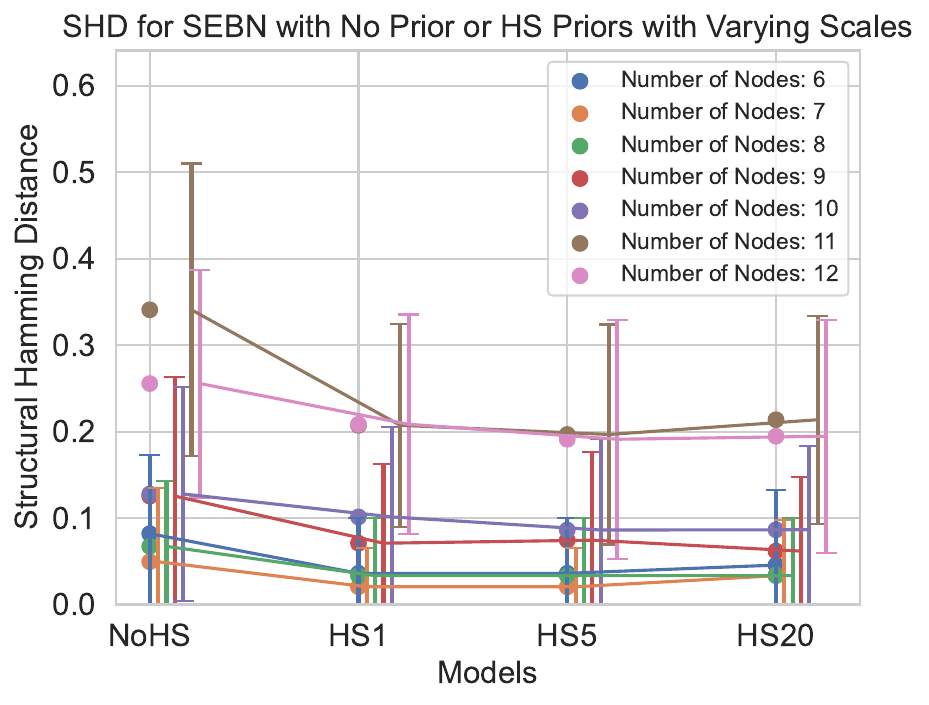}
            \caption{SHD vs. Number of Nodes for models without Horseshoe prior or with a Horseshoe prior scale $\tau = 1, 5, 20$, each with $w_HS = 1$.}
            \label{unpd:hs1hs20_shd}
        \end{subfigure}
        
        \begin{subfigure}{\linewidth}
            \includegraphics[width=\linewidth]{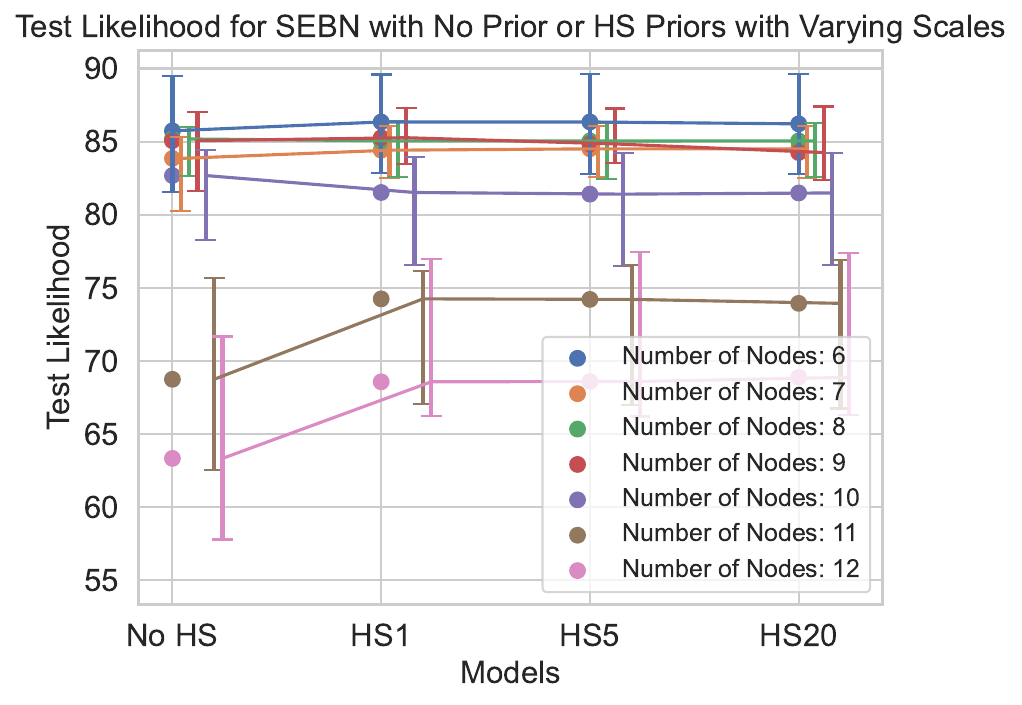}
            \captionsetup{width=0.9\linewidth}
            \caption{Test Likelihood vs. Number of Nodes for models without Horseshoe prior or with a Horseshoe prior scale $\tau = 1, 5, 20$, each with $w_HS = 1$.}
            \label{unpd:hs1hs20_lik}
        \end{subfigure}
    \end{minipage}%
    \begin{minipage}{0.4\textwidth}
        \centering
        \begin{subfigure}{\linewidth}
            \includegraphics[width=\linewidth]{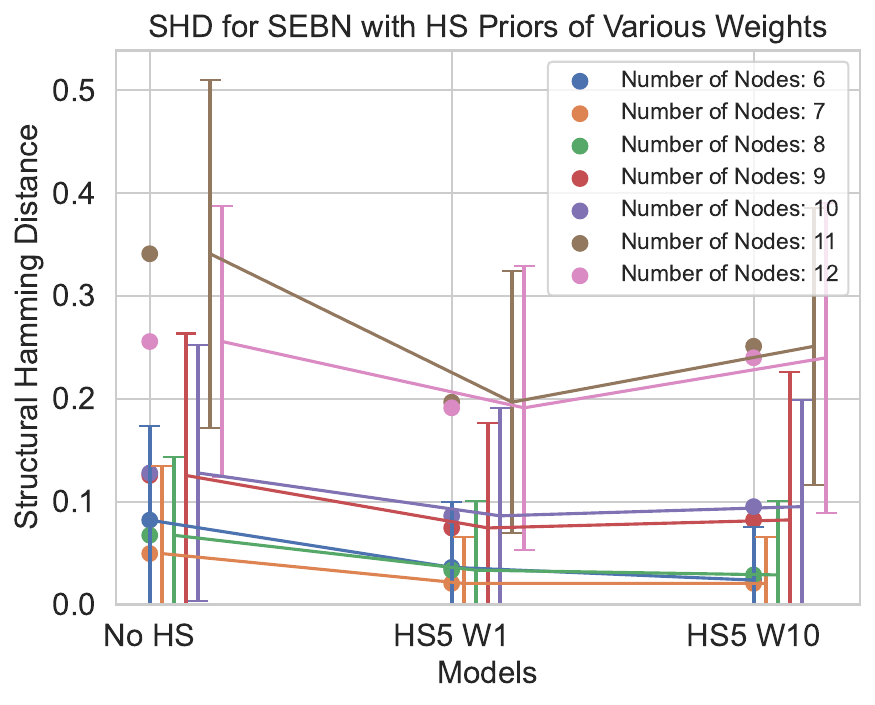}
            \captionsetup{width=0.9\linewidth}
            \caption{SHD vs. Number of Nodes for models with Horseshoe prior scale $\tau = 5$ , each with a weight $w_{HS} = 1, 10$.}
            \label{unpd:w1w10_shd}
        \end{subfigure}
        
        \begin{subfigure}{\linewidth}
            \includegraphics[width=\linewidth]{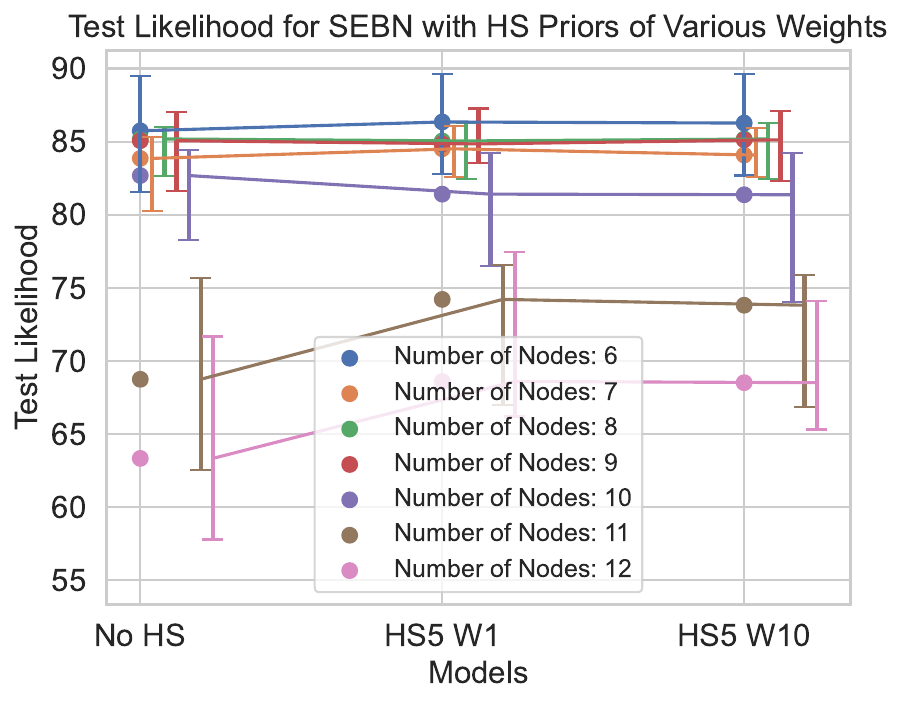}
            \captionsetup{width=0.9\linewidth}
            \caption{Test Likelihood vs. Number of Nodes for models with a weight $w_{HS} = 1$ or $w_{HS} = 10$, each with Horseshoe prior scale $\tau = 5$ }
            \label{unpd:w1w10_lik}
        \end{subfigure}
    \end{minipage}

    % FDV: What is the take-away?
    \caption{In Independent-addition dataset, (a) and (b): In SHD and Test likelihood, all models with the Horseshoe prior outperformed one without. The performance trends across different Horseshoe scales had a U-shaped pattern. Too small or large scales penalized GP edges too severely or lightly, resulting in incorrect structure. $\tau = 5$ emerged as the optimal choice. (c) and (d): In SHD and Test likelihood, $w_{HS} = 1$ consistently outperformed $w_{HS} = 10$.}
    \label{unpd}
\end{figure*}

\paragraph{Dataset Generation and Sampling} To compare the performance of the models, we generated 100 different synthetic Bayesian networks with $6 \leq n \leq 12$ continuous variables and at most $n \cdot (n-1)/2$ edges. For each Bayesian network, we sampled 500 instances for training, 100 instances for evaluation and another 100 for test.

\paragraph*{UCI Liver Disorders}

From the UCI Machine Learning Repository, we chose the Liver Disorders dataset \cite{misc_liver_disorders_60}, which comprises 345 samples of 7 continuous variables. One constant variable was excluded for analysis. The training and test samples were randomly split with a ratio of 9:1.

\paragraph{Baselines}

\subparagraph{Two-step Learning Baseline}
We compared the performance of determining linear components first and fixing them throughout optimizing GP components solely, as opposed to cotraining them.

Given Ground-Truth Linear Terms:
In this approach, we assumed domain experts provided the linear terms, eliminating the need to learn them from the data. The oracle ensures GP components only fit nonlinearity in the data, creating an ideal scenario to assess upper limit performance.

Fitted and Fixed Linear Terms:
In the second approach, we fitted the linear terms to the data and learned the GP based on the fitted and fixed linear terms. This allowed us to obtain the best possible linear terms for our dataset and learn a GP that captured the remaining non-linear trends. A potential issue is that the fitted linear terms may attempt to explain nonlinear components, hindering learning the true GP parameters.

\subparagraph{SPBN Baseline} 
SPBN-CKDE \cite{atienza2022semiparametric} learns nodes as either parametric (linear Gaussian, LG) or non-parametric (KDE) type. The parametric and nonparametric components are optimized together in a modified structure learning hill climbing algorithm with an extra operator -- node type change. The initial node type for all nodes is nonparametric, i.e., the CPDs are KDEs. The starting graph had no edges. Patience $\lambda$ is 15.

\paragraph{Optimization}
Expert graph is given as ground truth in synthetic dataset and learned using the Linear Gaussian Bayesian Network (LGBN) algorithm \cite{lauritzen1989graphical} in UCI dataset. Noise levels are assumed known for each variable. Initial values of GP amplitudes and GP length scales were set to 0.2 and 0.4, respectively, approximating the ground truth. To assessed the effect of the Horseshoe prior on GP and identify the optimal scale, we experimented with no prior, or a range of Horseshoe scales from 1 to 20. Additionally, we experimented with Horseshoe weights of 1 and 10 to determine the most effective regularization level.

A small GP amplitude suggests minimal non-parametric variation, indicating a potential lack of need for GP edges. Consequently, a lower limit was set. 
% FDV: How was the tuning done?
% Gwen: for synthetic, in models with no prior, inspect the learned amplitudes for true and false edges, take a few values in the middle (far from both ends) and grid search for the closest to ground truth. for real world data, same without the ground truth comparison.
% the reason for one single threshold for all HS scales is to keep one factor constant and only consider effect of scales, but from another perspective should we tune the best threshold for each scale? then we expect smaller optimal threshold for smaller scale, and overall less difference in effect of HS.
The optimal threshold was tuned to 0.2 for Independent-addition Dataset (ID), 0.1 for Expert-guided Dataset (ED), and 0.01 for UCI Liver Disorders.

We performed parameter optimization using GPyTorch. The maximum number of iterations is 200, with a patience parameter (\(\lambda\)) of 20. The best model is determined based on the highest likelihood in the validation set \(D_{\text{val}}\).

\paragraph{Metrics}
\subparagraph{Structural Hamming Distance (SHD):}
SHD \cite{tsamardinos2006max} measures the number of edge additions, removals, or reversals required to transform one graph into another. SHD results are reported as means.
\subparagraph{Test Likelihood:}
Test likelihood estimates the expected performance of the models on new and unseen data by calculating the log-likelihood of the test dataset. All "test likelihood" in the Results section are log-likelihood and reported as medians.

\section{Results}

\begin{figure*}[h]
\centering
    \begin{minipage}{0.4\textwidth}
        \centering
        \begin{subfigure}{\linewidth}
            \includegraphics[width=\linewidth]
            {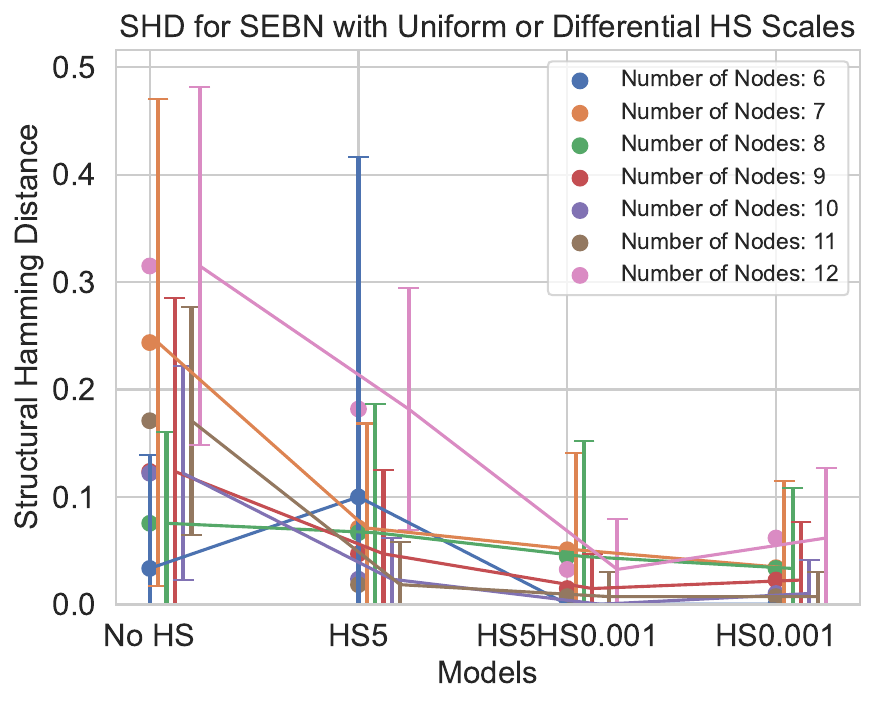}
            \captionsetup{width=0.9\linewidth}
            \caption{SHD vs. Number of Nodes for four models: one with no Horseshoe prior, one with a differential Horseshoe prior scales $\tau = 5$  for expert edges and $\tau = 0.001$  for non-expert edges, and another two with a uniform scale $\tau = 5$  or $\tau = 0.001$  for all possible edges.}
            \label{egned:shd}
        \end{subfigure}
    \end{minipage}%
    \begin{minipage}{0.4\textwidth}
        \centering
        \begin{subfigure}{\linewidth}
            \includegraphics[width=\linewidth]{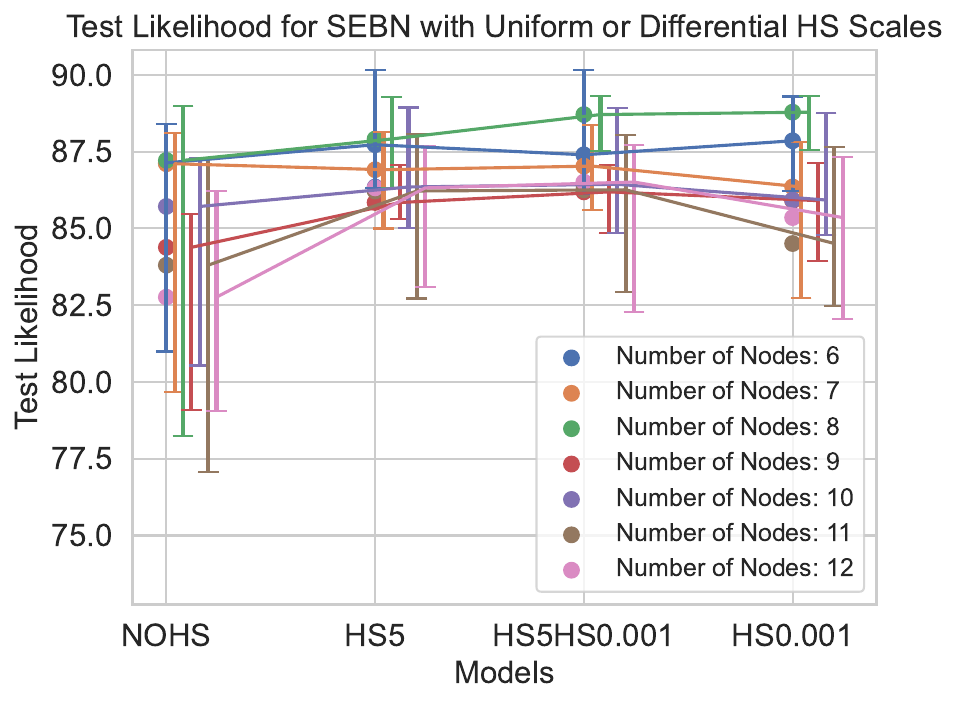}
            \captionsetup{width=0.9\linewidth}
            \caption{Test Likelihood vs. Number of Nodes for four models: one with no Horseshoe prior, one with a differential Horseshoe prior scales $\tau = 5$  for expert edges and $\tau = 0.001$  for non-expert edges, and another two with a uniform scale $\tau = 5$  or $\tau = 0.001$  for all possible edges.}
            \label{egned:lik}
        \end{subfigure}
    \end{minipage}

    % FDV: Write out the data set name 
    \caption{In Expert-guided dataset, Horseshoe prior improved SHD (a) and test likelihood (b) in most cases. The model with differential Horseshoe prior scales consistently outperforms those with uniform scales, be it too small ($\tau = 0.001$) or too large ($\tau = 5$ ).}
    \label{egned}
\end{figure*}

\paragraph{Finding Structure: In the independent-addition setting, our SEBN added a minimal number of GP edges across all network sizes.}
Figure~\ref{fig:baseline}(a) shows the Structural Hamming Distance (SHD) as the network size grows for our SEBN approach, the oracle with the ground-truth linear parameters (would not be available in real settings), and a two-stage process that fits the linear model before the nonlinear model.  Overall, SHD increased with the network size. At all network sizes, our jointly-trained approach got about 50\% closer to the oracle solution than the two-stage approach.  We used the jointly-trained approach for the remaining experiments.

The SEBN model in Figure~\ref{fig:baseline}(a) was not regularized by a Horseshoe prior.

Figure~\ref{unpd}(a) shows the effect of different Horseshoe prior scales on learning the true structure. All models with the Horseshoe prior demonstrated improved performance compared to the model without it. The performance trends across different Horseshoe scales had a U-shaped pattern. Too small or large scales penalized GP edges too severely or lightly, resulting in incorrect structure. For our independent addition dataset, a Horseshoe prior scale of 5 emerged as the optimal choice.  That said, the results were fairly stable across scales, suggesting that this parameter should be relatively easy to tune.
 
Additionally, the heavier regularization ($\tau$ = 1) yielded better performance in smaller networks. Conversely, in larger networks, lighter regularization ($\tau$ = 20) performed better, although results were somewhat mixed. A detailed analysis of individual learned models across various network sizes revealed that in smaller networks, our SEBN learned extra edges, potentially overfitting to the training data. In more complex networks, our SEBN learned a combination of extra and missing edges, and as a result, lighter regularization ($\tau$ = 20) showed relatively better performance. A suitable middle ground, achieved with $\tau = 5$, strikes a balance between extra and missing edges.

Besides scales, Fig. \ref{unpd}(c) showed the impact of different weights on the Horseshoe prior $w_{HS}$. $w_{HS} = 1$ consistently outperformed $w_{HS} = 10$, especially in larger networks. This result aligned with our earlier observations that in complex networks, lighter regularization showed better performance. Consequently, we used $w_{HS} = 1$ for all subsequent results.

Lastly, our SEBN, with or without Horseshoe prior, significantly outperformed the state-of-the-art baseline SPBN (Fig. \ref{fig:baseline}(c)). SEBN, but not SPBN, allows edges to contain both linear and GP components.  This feature aligns well with our synthetic data setting, which aims to closely mimic real-world scenarios.

\paragraph{In the independent-addition setting, our SEBN  predicts unseen data well.}
% FROM SYNTHETIC EXPERIMENTS ABOVE
% FIGURE/TABLE WITH TEST LIKELIHOOD AS VARIOUS FACTORS CHANGE

Figure~\ref{fig:baseline}(b) shows the test likelihood as the network size grows for our SEBN approach, the oracle, and a two-stage process.  Mirroring the SHD results, overall, test likelihood increased with the network size. At all network sizes, our jointly-trained approach got about 50\% closer to the oracle solution than the two-stage approach.

Similarly, Fig. \ref{unpd}(b) shows the predictive performance of different Horseshoe prior scales. The incorporation of Horseshoe prior improved test likelihood. The performances across various Horseshoe scales had an inverse U-shaped pattern, indicating suboptimal performance for both excessively small and large scales. Furthermore, the weight of Horseshoe prior $w_{HS} = 1$ consistently outperformed $w_{HS} = 10$ (Fig. \ref{unpd}(d)).

\paragraph{In expert-guided setting, our SEBN with differential Horseshoe scales correctly prioritizes modifying expert edges over adding extra edges and explains the data better.} 
% SYNTHETIC EXPERIMENT: AS VARIOUS FACTORS (DATA, NODES, EDGES) CHANGE, AS LONG AS THE EXPERT GRAPH AND THE DATA HAVE THE SAME EDGES, NO/FEW EDGES ARE ADDED COMPARED TO ALTERNATIVES (WHAT ARE THE ALTERNATIVES??) 

% FIGURE/TABLE WITH EDGES ADDED AS VARIOUS FACTORS CHANGE: design a presentation

For Expert-guided Dataset (ED), Fig. \ref{egned}(a) and Fig. \ref{egned}(b) show SHD and test likelihood for SEBN with no Horseshoe prior or a prior with a uniform scale $\tau = 5$ or $\tau = 0.001$, or differential scales $\tau = 5$ for expert edges and $\tau = 0.001$ for non-expert edges.

Horseshoe prior improved SHD and test likelihood in most cases. Further, the model with differential Horseshoe prior scales consistently outperforms those with uniform scales, be it too small ($\tau = 0.001$) or too large ($\tau = 5$ ).

The models ranked differently in SHD and Test Likelihood. The model with a uniform scale $\tau = 0.001$ performs relatively worse in Test Likelihood than SHD, indicating that large regularization prunes excessive edges but also pushes parameters down, resulting in an acceptable structure but incorrect parameters. Conversely, the model with a uniform scale $\tau = 5$ performs relatively worse in SHD than Test Likelihood, suggesting that small regularization tends to learn more edges, leading to an incorrect structure but a high test likelihood.

\label{UCI}
\begin{figure*}[h]
\centering
    \begin{minipage}{0.25\textwidth}
        \centering
        \begin{subfigure}{\linewidth}
            \includegraphics[width=\linewidth]{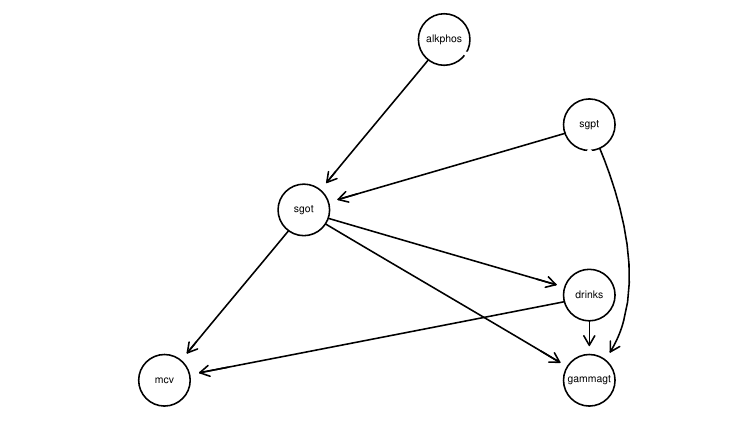}
            \caption{Expert graph}
            \label{UCI:expert}
        \end{subfigure}
    \end{minipage}%
    \begin{minipage}{0.25\textwidth}
        \centering
        \begin{subfigure}{\linewidth}
            \includegraphics[width=\linewidth]{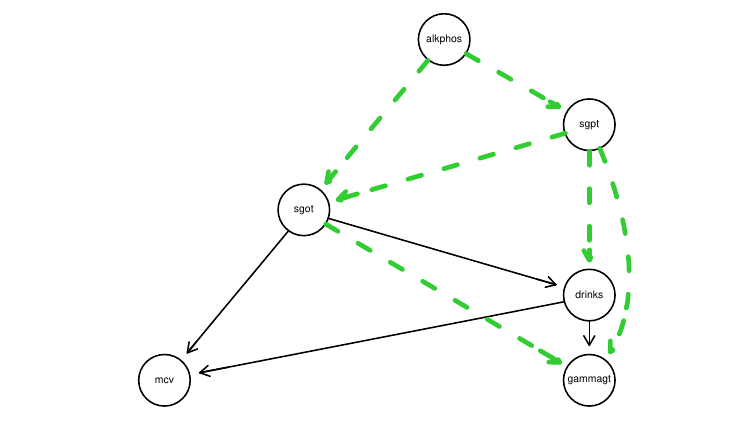}
            \caption{No Horseshoe}
            \label{UCI:nohs}
        \end{subfigure}
    \end{minipage}%
    \begin{minipage}{0.25\textwidth}
        \centering
        \begin{subfigure}{\linewidth}
            \includegraphics[width=\linewidth]{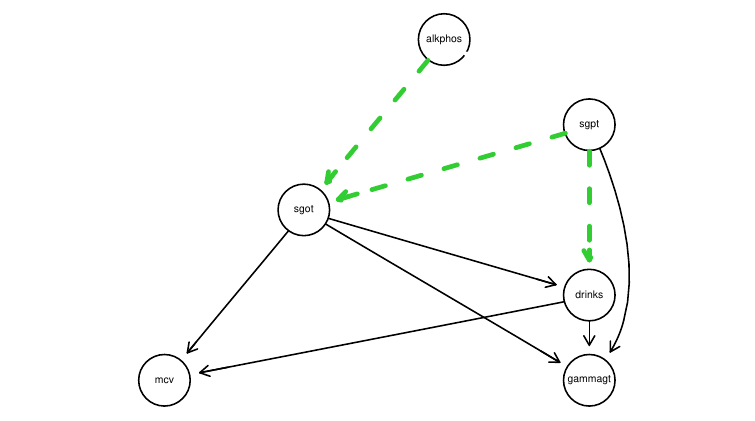}
            \caption{Uniform HS scale $\tau = 5$ }
            \label{UCI:hs5}
        \end{subfigure}
    \end{minipage}%
    \newline
    \begin{minipage}{0.25\textwidth}
        \centering
        \begin{subfigure}{\linewidth}
            \includegraphics[width=\linewidth]{figure/hs10_graph.pdf}
            \caption{Differential HS scales $\tau = 5$  and $\tau = 0.001$ }
            \label{UCI:hs5hs0.001}
        \end{subfigure}
    \end{minipage}%
    \begin{minipage}{0.25\textwidth}
        \centering
        \begin{subfigure}{\linewidth}
            \includegraphics[width=\linewidth]{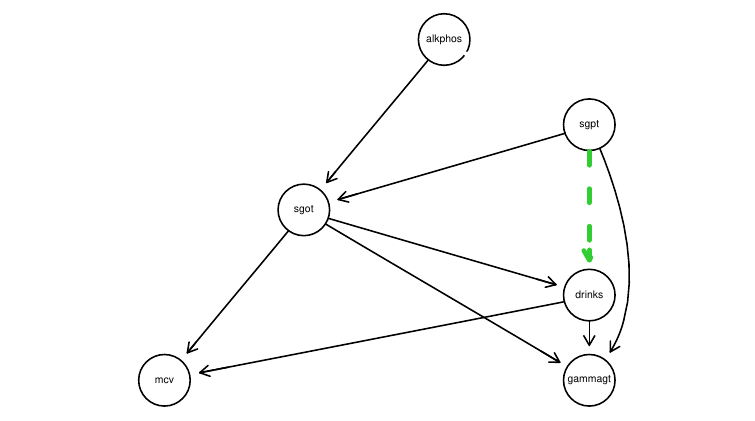}
            \caption{Uniform HS scale $\tau = 5$ }
            \label{UCI:hs0.001}
        \end{subfigure}
    \end{minipage}%
    \begin{minipage}{0.25\textwidth}
        \centering
        \begin{subfigure}{\linewidth}
            \includegraphics[width=\linewidth]{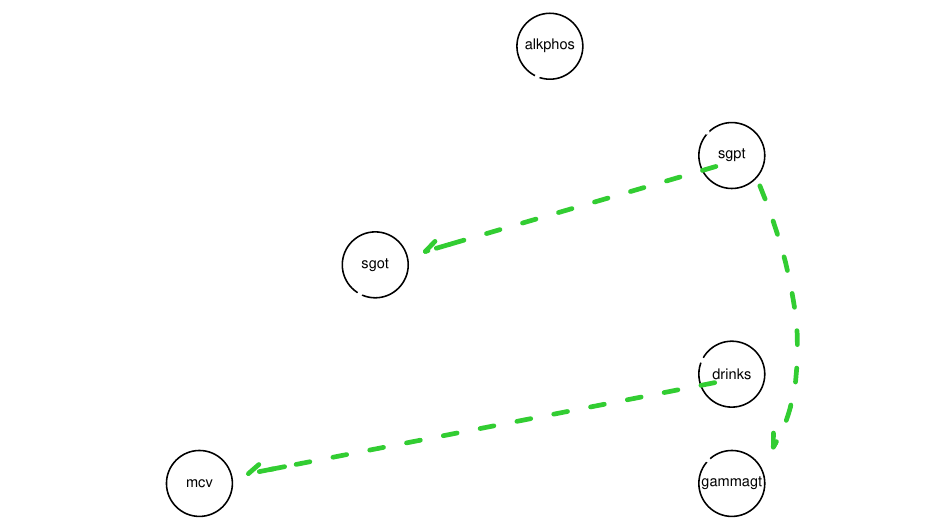}
            \caption{SPBN}
            \label{UCI:spbn_graph}
        \end{subfigure}
    \end{minipage}
    
    \caption{Learned graphs for UCI Liver Disorders: (a) Expert graph learned as a linear GBN. (b) Graph learned with No HS, incorporating 6 additional GP edges with a test likelihood of -72.31. (c) Uniform HS scale $\tau = 5$ , adding 3 GP edges and achieving a test log likelihood of -72.31. (d) Differential HS scales $\tau = 5$ and $\tau = 0.001$ for expert and non-expert graphs respectively, introducing 3 GP edges with a test likelihood of -60.56. (e) Uniform HS scale $\tau = 0.001$ , adding 1 GP edge with a test likelihood of -60.57. (f) SPBN, learning 4 non-parametric edges with a test likelihood of -213.06.}
    \label{UCI}
\end{figure*}

\paragraph{On UCI Liver Disorders Dataset, SEBN generated diverse graphs with varying Horseshoe scale settings. A smaller Horseshoe scale added fewer GP edges, and differential scales improved test likelihood. All SEBN models outperformed SPBN in test likelihood.}
For UCI Liver Disorders dataset, "expert graph" learned using the Linear Gaussian Bayesian Network (LGBN) algorithm \cite{lauritzen1989graphical} consists of six edges in the six-node BN (Fig. \ref{UCI}(a)). SEBN learned four candidate graphs with no Horseshoe prior (Fig. \ref{UCI}(b)), a uniform Horseshoe prior $\tau = 5$ (Fig. \ref{UCI}(c)) or $\tau = 0.001$ (Fig. \ref{UCI}(d)), and differential Horseshoe prior scales $\tau = 5$ for expert edges and $\tau = 0.001$ for non-expert edges (Fig. \ref{UCI}(e)).

The use of a uniform Horseshoe prior, and more significantly, one with a smaller scale, resulted in reduced additions of GP edges and improved test likelihood.

Differential Horseshoe prior scales further improved test likelihood. Although models with $\tau = 5$ and differential scales $\tau = 5$ or $\tau = 0.001$ learn the same number of edges, the latter had a higher test likelihood. Differential scales, which regularized parameters more strongly for non-expert edges, performed better in predicting unseen data.

The baseline model, SPBN, learned a structure significantly different from our models (Fig. \ref{UCI}(f)), as it contained non-parametric edges only. While a direct structural comparison lacks a common standard, SEBN consistently and significantly outperform the SPBN model in test likelihood (Table \ref{table_label}).

\begin{table}[h]
\centering
\begin{tabular}{l|l}
    \textbf{Model} & \textbf{Test Likelihood} \\
    \hline
    SPBN & -213.06 \\
    No HS & -72.31 \\
    HS 5 & -72.31 \\
    HS 5 + HS 0.001 & -60.56 \\
    HS 0.001 & -60.57 \\
\end{tabular}
\caption{Test likelihoods of SPBN and our models across Horseshoe scales.}
\label{table_label}
\end{table}

\section{Discussion and Conclusion}

We proposed a model for learning both linear and nonlinear edges within the Expert Bayesian Network framework. Specifically, we employed Gaussian Processes (GP) to capture non-parametric nuances in data that are challenging to specify by experts. GPs offer the advantage of having an amplitude parameter to indicate edge strength and allow for effective regularization, as well as the ability to independently learn hyperparameters for each parent.

To regularize GP amplitudes, we introduced a Horseshoe prior. Horseshoe prior, with its thick-tailed density distribution, can deactivate weak GP edges and retain significant ones. We optimized the Horseshoe prior parameters based on Structural Hamming Distance (SHD) and test likelihood in synthetic data. In real-world scenarios where ground truth is unknown, we produced a range of candidate learned graphs with varying regularization, providing enhanced interpretability.

Leveraging GP hyperparameter independence for each parent, we developed an exact structure learning algorithm based on Dynamic Programming. Our algorithm efficiently prunes the search space with the acyclic constraint derived from the partial topological order provided by experts.

Our approach has limitations. The exact search algorithm has relative high computational costs and hinders integration common packages that use approximate score methods. Future work will focus on enhancing computational efficiency and scalability.

Gaussian Processes offer promise in modeling non-parametric components with regularization, as their hyperparameters provide information on edge inclusion. While amplitude intuitively signals edge strength, the length scale can also serve as a threshold for detecting nonlinearity. A large length scale suggests linearity in the variable's scale and thus a lack of need for nonlinear components. A small length scale implies excessive wiggling, potentially overfitting the noise. Further theoretical proof and optimization of these thresholds are required.

Beyond the fully observed case, we intend to explore partially observed settings. Our goal is to recover the joint distribution over the observed variables with minimal changes to the expert graph, while maintaining computational efficiency.

Our model is a valuable and interpretable tool for healthcare professionals and researchers. It captures nuanced nonlinear relationships aligned with existing knowledge and offers flexibility in choosing from a spectrum of learned graphs. The incorporation of expert knowledge as both prior belief and posterior requirement makes it promising for advancing the application of artificial intelligence in healthcare and related domains.

\section*{Acknowledgements}
YW and FDV acknowledge support for this work from the National Science Foundation under Grant No. IIS-1750358.  Any opinions, findings, and conclusions or recommendations expressed in this material are those of the author(s) and do not necessarily reflect the views of the National Science Foundation.

\bibliography{aaai24}
\end{document}